\documentclass[10pt,twocolumn,letterpaper]{article}

\usepackage{wacv}
\usepackage{times}
\usepackage{epsfig}
\usepackage{graphicx}
\usepackage{amsmath}
\usepackage{amssymb}
\usepackage{gensymb}
\usepackage{multirow}
\usepackage{rotating}
\usepackage{fancyhdr}
\usepackage{latexsym}
\usepackage{url}
\usepackage{algorithm}
\usepackage{algorithmic}
\usepackage{bm}
\usepackage{stmaryrd}
\usepackage{dsfont}
\usepackage{epstopdf}
\usepackage{booktabs}
\usepackage{dashrule}

\usepackage{amsfonts}
\usepackage{mathrsfs}

\usepackage{enumitem}
\usepackage[accsupp]{axessibility}  

\floatname{algorithm}{Procedure}

\AppendGraphicsExtensions{.tiff}

\usepackage{caption}
\usepackage{subcaption}

\usepackage[pagebackref=true,breaklinks=true,letterpaper=true,colorlinks,bookmarks=false]{hyperref}

\wacvfinalcopy 

\ifwacvfinal
\def\assignedStartPage{1} 
\fi

\ifwacvfinal
\else
\fi

\begin{document}


\title{PPCD-GAN: Progressive Pruning and Class-Aware Distillation \\ for Large-Scale Conditional GANs Compression}

\author{Duc Minh Vo$^{1}$\\
{\tt\small vmduc@nlab.ci.i.u-tokyo.ac.jp}
\and
Akihiro Sugimoto$^{2}$\\
{\tt\small sugimoto@nii.ac.jp}
\and
Hideki Nakayama$^{1}$\\
{\tt\small nakayama@ci.i.u-tokyo.ac.jp}
\and 
$^{1}$ The University of Tokyo, Japan
\and
$^{2}$ National Institute of Informatics, Japan
}

\maketitle
\ifwacvfinal\thispagestyle{empty}\fi

\begin{abstract}
We push forward neural network compression research by exploiting a novel challenging task of large-scale conditional generative adversarial networks (GANs) compression.
To this end, we propose a gradually shrinking GAN (\textbf{PPCD-GAN}) by introducing progressive pruning residual block (PP-Res) and class-aware distillation.
The PP-Res is an extension of the conventional residual block where each convolutional layer is followed by a learnable mask layer to progressively prune network parameters as training proceeds.
The class-aware distillation, on the other hand, enhances the stability of training by transferring immense knowledge from a well-trained teacher model through instructive attention maps.
We train the pruning and distillation processes simultaneously on a well-known GAN architecture in an end-to-end manner.
After training, all redundant parameters as well as the mask layers are discarded, yielding a lighter network while retaining the performance.
We comprehensively illustrate, on ImageNet $128 \times 128$ dataset, PPCD-GAN reduces up to 5.2$\times$ (81\%) parameters against state-of-the-arts while keeping better performance.

\end{abstract}

\section{Introduction} \label{sec:introduction}

Scaling-up network parameters in GANs~\cite{brock2018large,han18self-attention,miyato2018spectral} has brought breakthroughs in large-scale conditional image generation.
Nevertheless, such scaled-up models~\cite{brock2018large,han18self-attention,miyato2018spectral} often require enormously expensive computational cost and memory, limiting their adoption to usage under real scenarios such as mobile devices.
In this paper, we aim to reduce the size of a conditional GAN model in terms of the number of parameters while attaining performance comparably to SOTAs~\cite{brock2018large,han18self-attention,miyato2018spectral}.
We state this novel problem as {\it large-scale conditional GANs compression}, which can be regarded as a new branch of neural network compression research~\cite{han2016deep}.
Different from image-to-image GANs compression~\cite{fu2020autogan,li2020gan}, our targeting class-conditional GANs have to deal with hundreds of classes at the same time.
This novel problem undoubtedly bridges the gap between academia and industry, yet it has not been well explored in the literature.

\begin{figure}[t]
	\centering
	\includegraphics[width=0.95\linewidth]{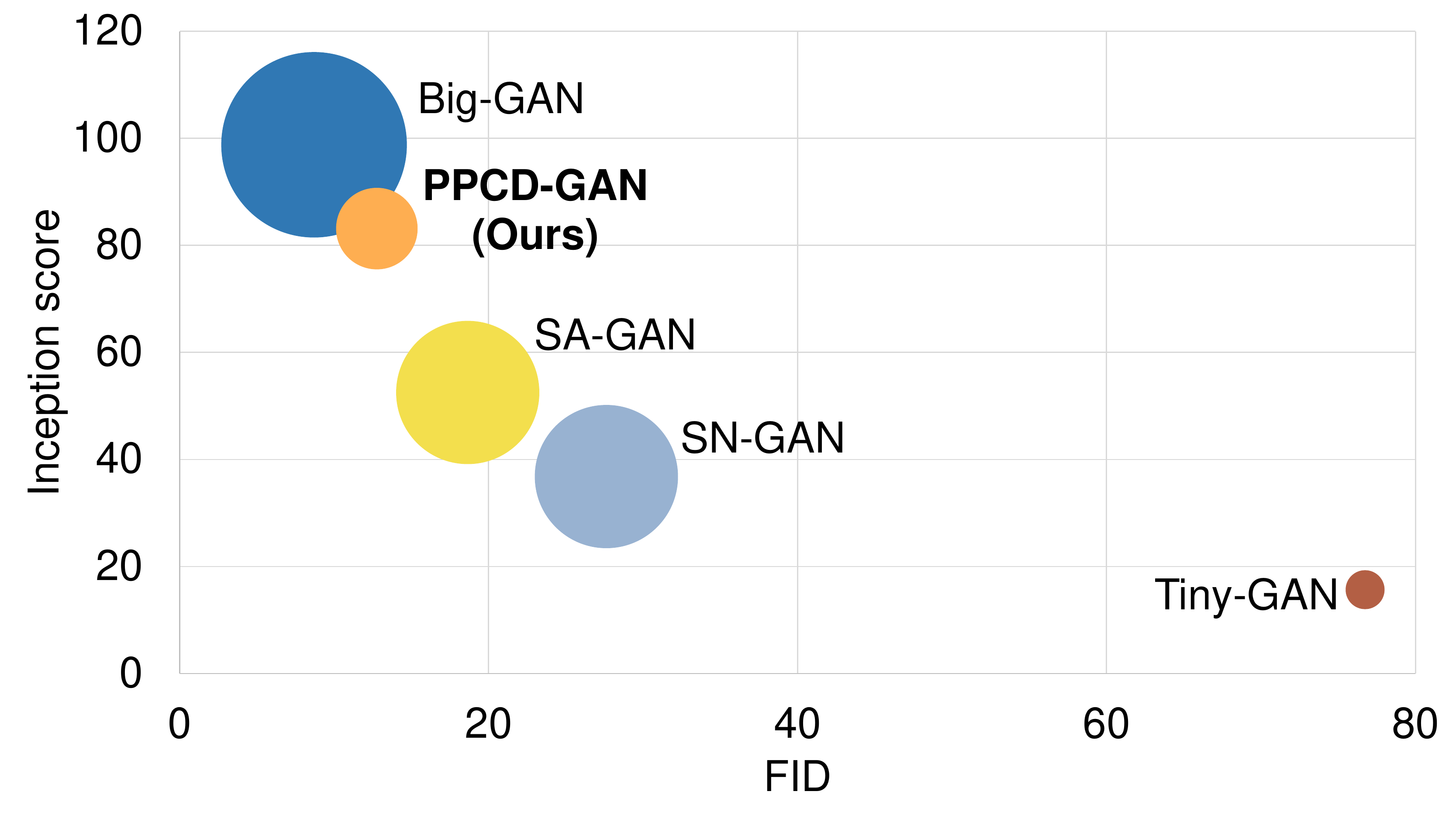}
			\vspace*{-0.5\baselineskip}
	\caption{Overall comparison on the number of parameters, Inception score (IS), and FID against Big-GAN~\cite{brock2018large}, SA-GAN~\cite{han18self-attention}, SN-GAN~\cite{miyato2018spectral}, and Tiny-GAN~\cite{chang2020tinygan}. PPCD-GAN is smaller than the others while achieving comparable IS and FID. The size of each circle reflects the number of parameters.} 
	\label{fig:overall_comparison}
	\vspace*{-\baselineskip}
\end{figure}

Neural network compression is long-standing research and most of its active works~\cite{han2016deep,han2015learning,luo2017thinet,frankle2019lottery,lin2020hrank,wang2020pruning,li2017pruning} rely on the two-step approach where the pruning step is followed by the re-training step.
The former step often directly removes network weights by employing pre-set criteria such as norm regularization~\cite{han2016deep,li2017pruning} or weight magnitude~\cite{han2015learning,frankle2019lottery}, producing a compressed model.
The latter one aims at restoring the performance of the compressed model with helps from fine-tuning or knowledge distillation~\cite{hinton2015distilling,zagoruyko2017AT}.
Although showing great potential in recognition and classification tasks, previous works cannot be straightforwardly extended to GANs compression~\cite{fu2020autogan,shu2019co-evolutionary} due to two inevitable reasons.
On one hand, the GANs model structure requires more parameters and more complicated computation than those of the typical convolutional neural networks (CNNs) model.
As a result, the GANs model is vulnerable to pre-set criteria, leading to drastically dropping its performance as addressed in~\cite{fu2020autogan,shu2019co-evolutionary}.
On the other hand, training GANs is notably unstable, and thus re-training the compressed model brings new challenges.
The above facts reveal difficulties in GANs compression.

Very recently, Fu et al.~\cite{fu2020autogan} and Li et al.~\cite{li2020gan} proposed a method for image-to-image GANs compression where neural architecture search (NAS)~\cite{elsken2019neural} on a trained GAN is used.
Li et al.~\cite{li2020gan} incorporated knowledge distillation~\cite{hinton2015distilling} to stablize the training.
The usage of NAS, however, requires a high computational cost due to a large search space.
Moreover, using NAS independently of the training process may break the underlying distribution of GANs because NAS itself does not know the distribution of a trained model.
To overcome these issues, the pruning and training processes are required to perform simultaneously.

Inspired by the above observations, we approach to our problem by unifying the pruning and distillation processes.
To this end, we introduce progressive pruning residual block (PP-Res) and class-aware distillation.
The PP-Res is mostly identical to the conventional residual block~\cite{he2016deep} except for that we add a learnable mask layer with its own parameters on top of each convolutional layer, leading to gradual removal of redundant weights as training proceeds.
All unnecessary weights and masks are safely removed after training without any risk.
The class-aware distillation, on the other hand, acquires knowledge from a pre-trained model (i.e., Big-GAN~\cite{brock2018large}) through attention maps~\cite{zagoruyko2017AT}, resulting in stability of training.
As Big-GAN~\cite{brock2018large} suffers from class leakage, we employ the class conditional normalization before computing the attention maps.
We simultaneously train the pruning (i.e., PP-Res) and distillation (i.e., class-aware distillation) processes in the end-to-end manner.
Our contributions are two-fold:

\begin{itemize}
    \item We address a novel problem of large-scale conditional GANs compression. While a compact model for conditional GANs was very recently proposed~\cite{chang2020tinygan}, it manually changes the architecture of Big-GAN~\cite{brock2018large} significantly, so that it successfully works on only homogeneous classes of a dataset (i.e., ImageNet~\cite{russakovsky2015imagenet} 398 animal classes). In contrast, to best of our knowledge, PPCD-GAN is the first GANs compression framework that fully explores a large-scale dataset.
    \item Our proposed \textit{PPCD-GAN} unifies pruning and distillation for the first time on the large-scale conditional GANs compression problem.
\end{itemize}

Comprehensive experiments on ImageNet $128 \times 128$ dataset~\cite{russakovsky2015imagenet} show the efficacy of PPCD-GAN against SOTAs (see Fig.~\ref{fig:overall_comparison} for comparison).

\section{Related Work}

\begin{figure*}[tb]
	\centering
	\includegraphics[width=0.8\linewidth]{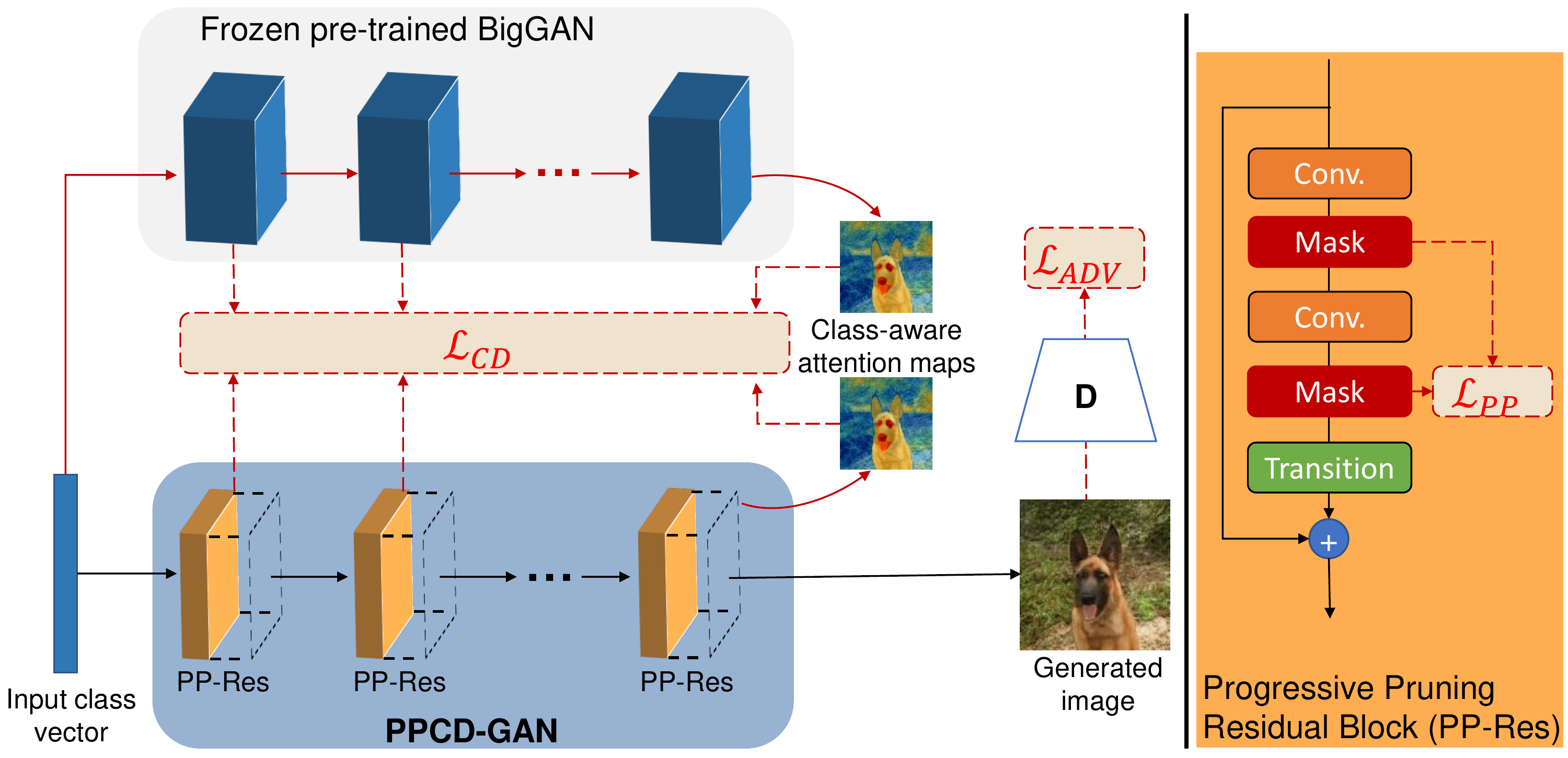}
	\caption{Overall framework of our proposed method (left) and progressive pruning residual block (right). For simplicity, we mainly illustrate important layers in the progressive pruning residual block. The red parts are removed after training.} 
	\label{fig:framework}
	\vspace*{-\baselineskip}
\end{figure*}

\noindent
\textbf{Large-scale conditional image generation.} Only a few methods~\cite{miyato2018spectral,han18self-attention,brock2018large} successfully handled this problem.
Miyato et al.~\cite{miyato2018spectral} used the label projection layer (i.e., SN-GAN) in the discriminator for better real/fake classification.
Zhang et al.~\cite{han18self-attention} incorporated a self-attention mechanism into GANs (i.e., SA-GAN), which allows the details of the image are generated by using cues from all feature locations.
Brock et al.~\cite{brock2018large} introduced Big-GAN, which is built upon SA-GAN~\cite{han18self-attention} by scaling up parameters and the batch size in training.
These methods require enormously expensive computational cost and memory. For instance, SN-GAN~\cite{miyato2018spectral} and SA-GAN~\cite{han18self-attention} models need $\sim$40M parameters whereas Big-GAN~\cite{brock2018large} requires even more ($\sim$70M).

Very recently, Chang et al.~\cite{chang2020tinygan} proposed Tiny-GAN with less parameters than~\cite{miyato2018spectral,han18self-attention,brock2018large}.
Tiny-GAN is a manually designed network based on the Big-GAN~\cite{brock2018large} architecture and replaces the standard convolutional layer with depth-wise separable convolution.
Moreover, it employs the generated images obtained by pre-trained Big-GAN~\cite{brock2018large} as the real images in training, which thus allows to distill black-box knowledge from Big-GAN~\cite{brock2018large}.
Despite such efforts, Tiny-GAN works properly only on a small sub-set of the large-scale ImageNet.
Different from Tiny-GAN~\cite{chang2020tinygan}, our method directly removes redundant parameters without significantly changing a well-known architecture.
Furthermore, we distill instructive knowledge from Big-GAN~\cite{brock2018large} through the attention maps, resulting in successfully handling the ImageNet in full-scale.

\noindent
\textbf{CNNs and GANs compression.} 
CNNs compression consists of three types~\cite{kang2020operation}: matrix decomposition, network quantization, and pruning.  Among them, the pruning approach actively receives more attentions~\cite{han2016deep,luo2017thinet,frankle2019lottery,lin2020hrank,wang2020pruning,li2017pruning}.
The current mainstream of pruning methods focuses on connection pruning~\cite{han2016deep,frankle2019lottery} and structure pruning~\cite{luo2017thinet,frankle2019lottery,lin2020hrank,wang2020pruning,li2017pruning}.
The connection pruning methods~\cite{han2016deep,frankle2019lottery} eliminate unnecessary connections using heuristics or optimization processes, which heavily depends on the hardware specification~\cite{kang2020operation}.
The structure pruning methods~\cite{luo2017thinet,lin2020hrank,wang2020pruning,li2017pruning} are, on the other hand, hardware-friendly and aim at reducing unnecessary channels or layers.
For instance, Li et al.~\cite{li2017pruning} employ $\ell_1$ norm to remove redundant channels and then fine-tune the pruned network to recover its performance.
Our method belongs to the structure pruning type where we remove superfluous channels.
However, we tackle the GANs compression problem where the network is more vulnerable than the typical CNNs~\cite{shu2019co-evolutionary,fu2020autogan}.

Some image-to-image translation GANs compression methods~\cite{shu2019co-evolutionary,fu2020autogan,li2020gan} have been proposed recently.
Shu et al.~\cite{shu2019co-evolutionary} proposed a co-evolution to regularize pruning on CycleGAN.
However, their method cannot be extended to other GANs architecture since it is bound to cycle-consistency loss and pre-trained discriminator.
Fu et al.~\cite{fu2020autogan} and Li et al.~\cite{li2020gan} concurrently employed NAS~\cite{elsken2019neural} in GANs compression.
Li et al.~\cite{li2020gan} further incorporated knowledge distillation~\cite{hinton2015distilling} for training stability.
However, NAS is not effective in a broad search space and using NAS in pruning separated from training may lead to the corruption of a pre-trained generator's underlying distribution.

Very recently, Kang et al.~\cite{kang2020operation} proposed to train a differentiable mask for each output channel (in a convolutional layer) to softly discard unnecessary channels without extra re-training.
Their differentiable mask is estimated based on the batch normalization and ReLU layers.
Inspired by~\cite{kang2020operation}, we also employ a learnable mask to remove useless channels gradually.
Yet, our mask has its own parameters, holding the advantage of the class conditional normalization.

\noindent
\textbf{Knowledge distillation.} This is widely used to transfer knowledge from a well-trained teacher model to a (compact) student model in order to enhance the performance of the student model~\cite{hinton2015distilling,zagoruyko2017AT}.
Existing methods~\cite{li2020gan,chang2020tinygan} showed the advantages of knowledge distillation in GANs compression.
We thus employ knowledge distillation for better training our network.
Different from the methods~\cite{li2020gan,chang2020tinygan}, we follow Zaroruyko et al.~\cite{zagoruyko2017AT} to use attention maps as instructive guidance.
This is because together with the class conditional normalization, the attention maps enable us to learn useful class-aware information from the teacher model.

\section{Proposed PPCD-GAN}

\subsection{Design of PPCD-GAN}

We designed PPCD-GAN in the context of unification of pruning and distillation.
We propose a tailored framework for efficient GANs, inspired by the student—teacher framework~\cite{hinton2015distilling}, because neither NAS~\cite{fu2020autogan,li2020gan} nor significant changes in a network architecture~\cite{chang2020tinygan} are applicable in compressing GANs.
Unlike~\cite{hinton2015distilling}, we aim to gradually prune network parameters as training proceeds while stabilizing the training with help of knowledge on distillation.
Therefore, our framework requires two conditions: (i) a robust pre-trained teacher model, and (ii) the architecture of the teacher model roughly agrees with that of the student model.
We remark that both the models should have the capability of handling large-scale conditional image generation.
Here, as an example, we choose SA-GAN~\cite{han18self-attention} as our student model whereas Big-GAN~\cite{brock2018large} as the teacher model because Big-GAN~\cite{brock2018large} is an extension of SA-GAN~\cite{han18self-attention} and achieves leader-board IS and FID.
Note that our framework can be applied to other GANs architectures as long as the above requirements are satisfied.

Residual blocks~\cite{he2016deep} are essential ingredients of SA-GAN~\cite{han18self-attention} and their parameters are a tremendous burden to the network parameters.
Accordingly, we pay our attention to reducing the parameters of the residual blocks, pruning the output channels (i.e., the number of convolutional filters) of convolutional layers.
Needless to say, directly cutting off several output channels is not effective~\cite{fu2020autogan,li2020gan}.
We thus follow the idea proposed in~\cite{kang2020operation} to progressively prune convolutional layers along with training.
In our case, because one residual block is connected to its previous block and/or the next one, pruning a residual block may cause channel inconsistency among different residual blocks.
To resolve this inconsistency, we introduce a transition layer on top of the residual block.
The transition layer ensures the channel consistency and efficiently works to avoid an explosion of the network parameters in general.

Since training GANs is notably unstable, we distill immense knowledge from a teacher model to settle down the training process.
We thus introduce the class-aware distillation to better transfer prior knowledge from Big-GAN to our network.
Fig.~\ref{fig:framework} (left) shows the PPCD-GAN framework.

Like SA-GAN~\cite{han18self-attention}, the generator of PPCD-GAN receives a pair of a noise vector and a class condition vector as its inputs.
The discriminator, on the other hand, receives a pair of an image and its corresponding class condition vector where the image can be either real or fake.

\subsection{Progressive pruning residual block}

Our progressive pruning residual block is based on the convention residual block~\cite{he2016deep}, but has two distinctions.
First, the \textit{learnable mask layer} enables the network to gradually remove channels during training while retaining the benefit of class conditional normalization (the key idea of conditional GANs).
Second, the \textit{transition layer} maintains the channel consistency across multiple residual blocks.

\noindent
\textbf{Learnable mask layer.} Inspired by~\cite{kang2020operation}, we place our learnable mask layer before the ReLU layer because negative activations are zeroed out by the ReLU layer. 
This realizes network parameters removal with low risk.
Although~\cite{kang2020operation} estimates the mask value based on the parameters of the batch normalization layer using a pre-defined constant, we build a mask layer with its own parameters.
This is because our model employs the conditional batch normalization where the class condition vector (obtained through an embeddings layer) is used to distinguish various classes.
We enforce the parameters of the batch normalization layer towards a pre-defined constant, so that the condition vectors fall inside a constant range.
As a result, the model may ignore the contribution of the class condition.
We, in what follows, introduce the design of the learnable mask layer and its loss term.

Let $n$ be output channels (i.e., the number of filters) of a convolutional layer.
We devise a mask $\mathbf{m}=\{m_1, m_2,...,m_n\}$ ($m_i \in [0, 1]$) from its learnable parameters $\mathbf{W}=\{w_1, w_2,...,w_n\}$.
Each value $m_i$ is estimated based on its corresponding parameter $w_i$.
To jointly optimize the mask layer and other layers through a simple gradient-based optimization, we employ a popular differentiable function (i.e., \textit{sigmoid}) with a constant $\delta$ for continuous relaxation as:
\begin{equation}
    m_i = \frac{1}{1 + \exp(-\delta w_i)}.
    \label{eq:m_i}
\end{equation}

Like~\cite{kang2020operation}, we impose sparse regularization $\ell_{\scriptscriptstyle{\rm PP}}$ in training the mask layer to progressively enforce its each value towards zero:
\begin{equation}
    \ell_{\scriptscriptstyle{\rm PP}} = \sum_{i=1}^{n}|w_i + 1|. 
    \label{eq:mask loss}
\end{equation}

Our learnable mask is multiplied with the outputs of its corresponding convolutional layer.
As training proceeds, the values of the mask gradually become zeros, meaning that the gradient to its corresponding convolutional layer decreases to zeros.
Consequently, the mask hides redundant weights (in a convolutional layer) during training.
Therefore, we can safely remove unnecessary weights when testing without any risk.
Note that $m_i$ is expected to become 0 or 1 after training, but the training fact shows that the expectation is not always satisfied.

We employ a ``binarization trick" to enforce $m_i$ to be 0 or 1.
Letting $c$ be the number of elements of $\{m_1, m_2,...,m_n\}$ whose value is less than or equal to a pivot (we choose 0.005 in our experiments), we define a $ratio = \frac{c}{n}$.
We note that the above pivot can be any sufficiently small value between 0 and 1 which is independent of training settings.
If $ratio$ is greater than a (given) compression ratio threshold $\alpha$, we set all $m_i$'s greater than 0.005 to 1, otherwise 0.
We then stop update the mask.
In this way, our learnable mask fills the gap between the continuous space in training and the discrete space in testing.
Our "binarization trick" can be summarized as follows:
\begin{equation}
    m_{i}^{*} = 
    \begin{cases}
    m_i \quad \text{if}\ ratio \leq \alpha \\
    1 \quad \text{if}\ m_i > 0.005\ \text{and}\ ratio > \alpha \\ 
    0 \quad \text{if}\ m_i \leq 0.005\ \text{and}\ ratio > \alpha
    \end{cases}.
    \label{eq:binarization}
\end{equation}

\noindent
\textbf{Transition layer.} When we prune a residual block, channel inconsistency among different residual blocks inevitably happens because two consecutive blocks are connected through residual connection~\cite{he2016deep}.
We thus propose a simple yet effective solution by adding the $1 \times 1$ convolution layer, called transition layer, before executing residual connection.
The usage of the transition layer is insightful because of two reasons.
First, it does not significantly affect the feature maps from the previous layer since the $1 \times 1$ convolution layer is widely used to increase/decrease output channels.
Second, its number of parameters is small enough not to cause the explosion of the network parameters.

Leveraging the learnable mask layer and the transition layer, we propose a progressive pruning residual block (PP-Res) to replace the commonly used residual block in SA-GAN~\cite{han18self-attention}.
More precisely, we incorporate the learnable mask layer on top of each convolutional layer and the transition layer on top of each residual block.
In the end, each residual block consists of two mask layers and one transition layer (Fig.~\ref{fig:framework} right).
Extending the learnable mask layer and the transition layer to multiple residual blocks is straightforward.
It is worth noting that our PP-Res does not significantly change the architecture of SA-GAN~\cite{han18self-attention}.

\subsection{Class-aware distillation}

To stabilize the training process, we transfer immense knowledge from a teacher model to our PPCD-GAN through the distillation process~\cite{hinton2015distilling}.
Though our teacher model (i.e., Big-GAN~\cite{brock2018large}) is powerful, we potentially have two issues to solve.
First, since the number of the output channels of Big-GAN is much larger than that of PPCD-GAN, we may fail in directly computing the norm distance (e.g., $\ell_1$).
Second, Big-GAN (sometimes) suffers from class leakage, which may affect the distilled knowledge.
A simple way is to adopt the learnable remapping~\cite{li2020gan}.
That solution, however, is not good in our case because it computes loss on a set of consecutive feature maps under the assumption on the set being a portion of the compressed (pruned) model.
We thus follow Zaroruyko et al.~\cite{zagoruyko2017AT} to use attention maps.
This is because the attention maps contain more instructive information.
More precisely, we extract two attention maps from the output features of the residual block (in Big-GAN~\cite{brock2018large}) and the output features of PP-Res (in our model).
Next, we compute the distance between the two attention maps.
To avoid class leakage, we apply the class conditional normalization on output features of Big-GAN before computing the attention maps.
Our class-aware distillation is rational because summing along all feature maps ignores the contributions of removed channels, and gives, in return, more informative feedback to remaining ones.

Let $\mathbf{O}^{\rm T}$ and $\mathbf{O}^{\rm S}$ be the (normalized) features of Big-GAN with the size of $C^{\rm T} \times H \times W$ and the features of PPCD-GAN with the size of $C^{\rm S} \times H \times W$ ($C^{\rm S}$ is smaller than $C^{\rm T}$).
Following~\cite{zagoruyko2017AT}, we compute the attention map of Big-GAN by $F^{\rm T} = \sum_{i=1}^{C^{\rm T}}|\mathbf{O}^{\rm T}(i,:,:)|^2$ (using Python notation) and that of PPCD-GAN by $F^{\rm S} = \sum_{i=1}^{C^{\rm S}}|\mathbf{O}^{\rm S}(i,:,:)|^2$.
We then formulate our class-aware distillation as:
\begin{equation}
    \ell_{\scriptscriptstyle{\rm CD}} = \left \| \frac{F^{\rm T}}{\left \| F^{\rm T} \right \|_2} - \frac{F^{\rm S}}{\left \| F^{\rm S} \right \|_2} \right \|_2. 
    \label{eq:class-aware distillation}
\end{equation}

\subsection{Loss function}

PPCD-GAN replaces all residual blocks of SA-GAN~\cite{han18self-attention} with PP-Res(es).
Therefore, Eq.~\eqref{eq:mask loss} is naturally aggregated to the progressive pruning loss:
\begin{equation}
    \mathcal{L}_{\rm PP} = \frac{1}{2N}\sum_{k=1}^{2N}\ell_{\scriptscriptstyle{\rm PP}_k}, \label{eq:lossPP}
\end{equation}
where $N$ is the number of PP-Res(es). $N$ is set to 5 in experiments.

We execute class-aware distillation (Eq.~\eqref{eq:class-aware distillation}) at some blocks.
We thus define the class-aware distillation loss as:
\begin{equation}
    \mathcal{L}_{\rm CD} = \frac{1}{L}\sum_{k=1}^{L}\ell_{\scriptscriptstyle{\rm CD}_k}, \label{eq:lossCD}
\end{equation}
where $L$ is the number of (residual) blocks to which we execute class-aware distillation.
In experiments, we set $L=4$ because we execute the distillation from the second to the fifth (residual) blocks.

We also employ adversarial loss:
\begin{eqnarray}
\label{eq:adversarial}
    \mathcal{L}_{\rm ADV} =  \underset{G^{\rm S}}{\min} \; \underset{D^{\rm S}}{\max} \: \mathbb{E}_{(I_{\rm real}, cls) \sim p_{\rm data}}\left [\log(D^{\rm S}(I_{\rm real},cls))  \right ]  \nonumber
    \\
    \qquad + \; \mathbb{E}_{cls \sim p_{\rm data}, z \sim p_{\rm z}}\left [\log(1 - D^{\rm S}(G^{\rm S}(z, cls), cls))  \right ],
\end{eqnarray}
where $G^{\rm S}$ and $D^{\rm S}$ are the generator and discriminator; $I_{\rm real}$ is a real image and $cls$ is the class condition sampled from all the training dataset $p_{\rm data}$. Similarly, $z$ is the noise vector sampled from Gaussian distribution $p_{\rm z}$. $\mathbb{E}(\cdot)$ denotes expectation over either $p_{\rm data}$ or $p_{\rm z}$.

The total loss is defined as:

\begin{equation}
    \mathcal{L} = 0.01\times\mathcal{L}_{\rm PP} + \mathcal{L}_{\rm CD} + \mathcal{L}_{\rm ADV}. \label{eq:loss function}
\end{equation}

We set an empirically determined small hyperparameter for $\mathcal{L}_{\rm PP}$ to prevent quick convergence of mask layers.
We minimize Eq.~\eqref{eq:loss function} in the end-to-end manner. The details of training procedure can be found in the supplementary.

\section{Experiments}
\vspace*{-0.1\baselineskip}

\begin{figure*}[tb]
	\centering
	\includegraphics[width=1\linewidth]{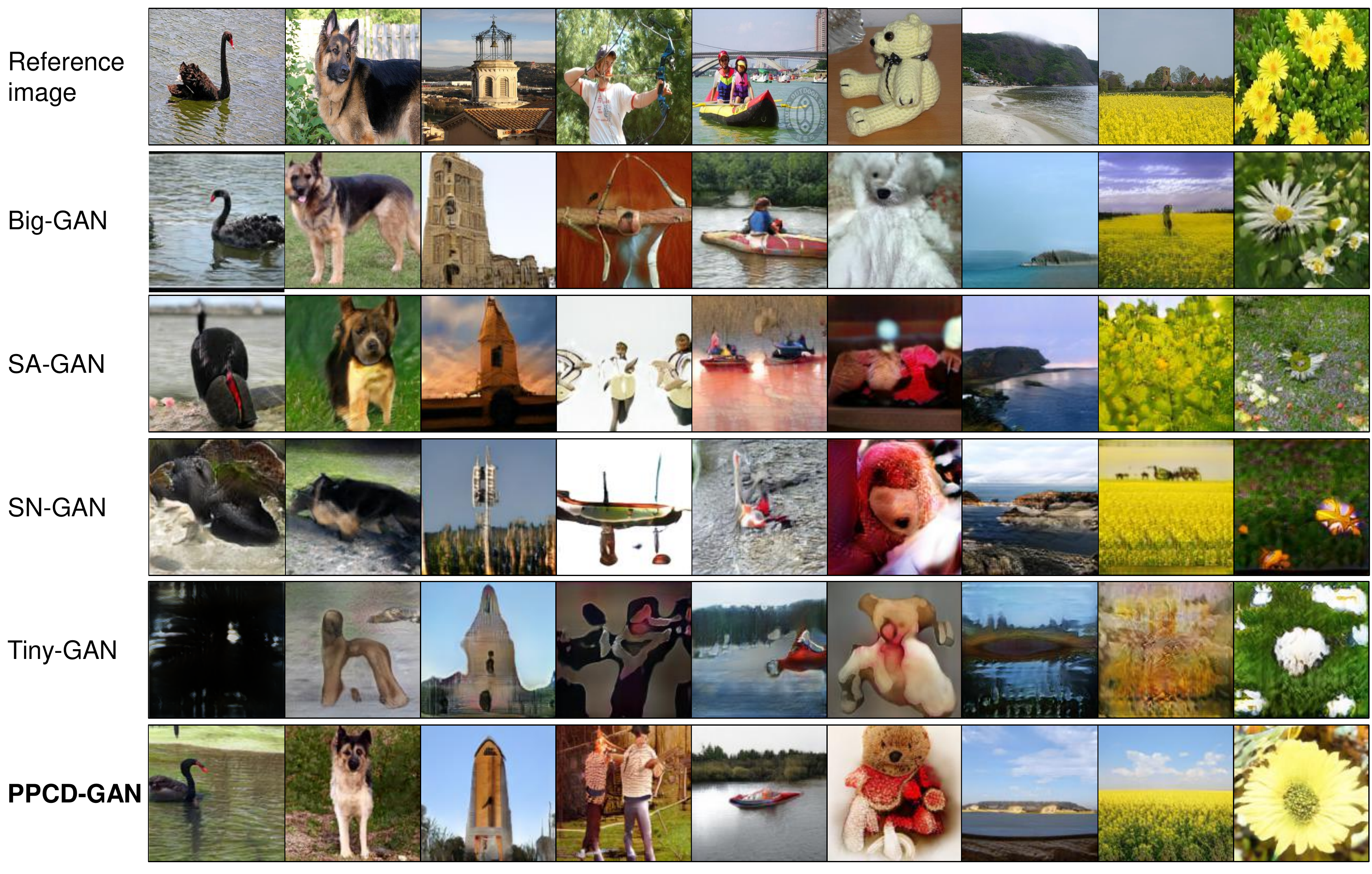}
			\vspace*{-1.5\baselineskip}
	\caption{Visual comparison of PPCD-GAN against Big-GAN~\cite{brock2018large}, SA-GAN~\cite{han18self-attention}, SN-GAN~\cite{miyato2018spectral}, and Tiny-GAN~\cite{chang2020tinygan}. From left to right, conditional classes are black swan, German shepherd, bell cote, bow, canoe, teddy, promontory, rapeseed, and daisy.} 
	\label{fig:visualization}
	\vspace*{-\baselineskip}
\end{figure*}

\subsection{Compared methods and evaluation metrics}
\vspace*{-0.12\baselineskip}

\noindent
\textbf{Dataset and compared methods.} 
We principally evaluated PPCD-GAN on a challenging large-scale ImageNet dataset~\cite{russakovsky2015imagenet}.
We note that although CIFAR and/or JFT-300 might be considered to be evaluated,
CIFAR is insufficient in scale (100 classes with size of $32 \times 32$) and JFT-300 is an undisclosed and internal dataset of Google.

We employ SA-GAN~\cite{han18self-attention} as the baseline.
We compare PPCD-GAN with Big-GAN~\cite{brock2018large}, SN-GAN~\cite{miyato2018spectral}, and Tiny-GAN~\cite{chang2020tinygan}.
For Big-GAN~\cite{brock2018large}, we employ pre-trained model released by the authors~\cite{biggan}.
We train SA-GAN~\cite{han18self-attention} on the re-implementation provided by Big-GAN's authors~\cite{biggan} since it produces better images.
For SN-GAN~\cite{miyato2018spectral} and Tiny-GAN~\cite{chang2020tinygan}, we trained authors' provided codes (SN-GAN~\cite{sngan}, Tiny-GAN~\cite{tinygan}) on ImageNet.
All networks are trained with their original settings.

Taking into account the ability of the compared models
and our computing resource, all the models are designed to generate images with the size of $3 \times 128 \times 128$.

\noindent
\textbf{Evaluation metrics.} We evaluate the overall quality of generated images using Inception score (IS)~\cite{salimans2016improved}, Fr\'{e}chet Inception Distance (FID)~\cite{heusel17GANs}, and diversity score (LPIPS)~\cite{zhang2018perceptual} (implemented in~\cite{IS,FID,LPIPS}).
We also report the number of parameters (\#Par.) and MACs~\cite{jebashini2015survey} to intuitively compare the compression rates.
We compute \#Par. and MACs using torchprofile tool~\cite{torchprofile}.
To compare practical speed of all compared methods, we measure the wall clock generating time on GPU (NVIDIA GTX3090) and CPU (AMD Ryzen 9 3900X) with the batch size of 64.

\subsection{Implementation and training details}

PPCD-GAN was implemented using PyTorch~\cite{PyTorch}.
The network was trained for 100 epochs using Adam optimizer~\cite{kingma2014adam}.
We set the initial learning rate of 0.1, which was dropped by a factor of 10 at the 30th, 60th, and 90th epochs.
In our experiments, the network was with the batch size of 128, and 2 gradient accumulations (equivalent to the batch size of 256, refer~\cite{biggan} for more details).  
Empirically, we set $\delta=10^3$ (Eq.~\eqref{eq:m_i}) and $\alpha=0.7$ (Eq.~\eqref{eq:binarization}).
All the networks were trained from scratch on a PC with GTX3090$\times$2 where
the parameters of the teacher model (i.e., Big-GAN) were fixed.
It took about 10 days to train each model.

\subsection{Comparison with state-of-the-arts}

\begin{table*}[tb]
\centering
\caption{Quantitative comparison against other conditional GANs methods (the best in \textcolor[rgb]{0,0,1}{blue}; the second best in \textcolor[rgb]{1,0,0}{red}).} 
	\vspace*{-0.75\baselineskip}
\label{tab:quantitative_comparison}
\begin{tabular}{l|ccc|cc|cc}
\midrule
\multirow{2}{*}{Models} & \multicolumn{3}{c}{\textbf{Quality}} &  \multicolumn{2}{c}{\textbf{Complexity}} &  \multicolumn{2}{c}{\textbf{Runtime} (second/batch)} \\
 & IS $\uparrow$ &  FID $\downarrow$  & LPIPS $\uparrow$ & \#Par. $\downarrow$ & MACs $\downarrow$ &  GPU $\downarrow$ & CPU $\downarrow$ \\
\midrule
Big-GAN~\cite{brock2018large} & \textcolor[rgb]{0,0,1}{98.8} & \textcolor[rgb]{0,0,1}{8.70} & 0.58 & 70.4M (1.0$\times$) & 21.3B (1.0$\times$) & 2.42 & 4.59 \\
SA-GAN~\cite{han18self-attention} & 52.52 & 18.65 & \textcolor[rgb]{0,0,1}{0.63} & 42M (1.7$\times$) & 9.6B (2.2$\times$) & 1.63 & 2.65 \\
SN-GAN~\cite{miyato2018spectral} & 36.80 & 27.62 & \textcolor[rgb]{1,0,0}{0.62} & 42M (1.7$\times$) & 9.1B (2.2$\times$) & 1.47 &\textcolor[rgb]{1,0,0}{2.35} \\
Tiny-GAN~\cite{chang2020tinygan} & 15.66 & 76.74 & 0.40 & \textcolor[rgb]{0,0,1}{3.1M} (22.7$\times$) & \textcolor[rgb]{0,0,1}{0.44B} (48.4$\times$) & \textcolor[rgb]{0,0,1}{0.45} & 3.62 \\
\textbf{PPCD-GAN} ($\alpha=0.7$) & \textcolor[rgb]{1,0,0}{83.13} & \textcolor[rgb]{1,0,0}{12.76} & \textcolor[rgb]{1,0,0}{0.62} & \textcolor[rgb]{1,0,0}{13.6M} (5.2$\times$) & \textcolor[rgb]{1,0,0}{1.6B} (13.3$\times$) & \textcolor[rgb]{1,0,0}{1.19} & \textcolor[rgb]{0,0,1}{2.05} \\
\midrule
\end{tabular}
\vspace*{-\baselineskip}
\end{table*}

\noindent
\textbf{Qualitative evaluation.} Fig.~\ref{fig:visualization} shows randomly selected examples of generated images obtained by PPCD-GAN and SOTAs~\cite{brock2018large,han18self-attention,miyato2018spectral,chang2020tinygan}.
Since each method receives a different noise vector as its input (along with the class vector), the generated images become different.
We thus run each method 10 times for a fair comparison, and picked up the best generated images based on the Inception score.
We see that the visual impressions of generated images by our method are on a par with those by Big-GAN~\cite{brock2018large}, and much better than those by the other methods (SA-GAN~\cite{han18self-attention}, SN-GAN~\cite{miyato2018spectral}, and Tiny-GAN~\cite{chang2020tinygan}).
SA-GAN~\cite{han18self-attention} is able to generate reasonable images, but the generated background is sometimes not clear (see the 2nd and 5th columns, for examples).
The generated images by SN-GAN~\cite{miyato2018spectral} regarding natural scenes are reasonable (see the 7th and 8th columns).
However, we observe low quality in generating objects with more details.
For Tiny-GAN~\cite{tinygan}, the generated images are incredibly miserable.
Interestingly, we visually observe that Big-GAN~\cite{brock2018large} generates images with class leakage (see the images in the 6th and 8th columns) while PPCD-GAN is not the case.
We attribute this advance to our usage of the class conditional normalization in the distillation process.

\noindent
\textbf{Quantitative evaluation.} First, we evaluated the overall quality of generated images (Table~\ref{tab:quantitative_comparison}, the second block).
We see that PPCD-GAN significantly outperforms SA-GAN~\cite{han18self-attention}, SN-GAN~\cite{miyato2018spectral}, and Tiny-GAN~\cite{chang2020tinygan} on IS and FID while achieving comparable levels with Big-GAN~\cite{brock2018large}.
This is because PPCD-GAN learns the immense knowledge from Big-GAN~\cite{brock2018large} thanks to our class-aware distillation.
Moreover, PPCD-GAN, SA-GAN~\cite{han18self-attention}, and SN-GAN~\cite{miyato2018spectral} are all able to achieve diversity (see LPIPS scores; we report the average over all the classes) whereas
Big-GAN~\cite{brock2018large} and Tiny-GAN~\cite{chang2020tinygan} do less diversity.
This less diversity comes from their usage of a ``truncation trick" which aims at the trade-off between variety and fidelity.
Note that LPIPS of each model is computed independently of the others; although the noise vector varies for each model, LPIPS always holds true.
We may conclude that PPCD-GAN is more effective in perceptive visuals and diversity than the others.

Next, we evaluated the complexity of each model using \#Par. and MACs (Table~\ref{tab:quantitative_comparison}, the third block).
We observe that PPCD-GAN significantly reduces the complexity.
More precisely, when compared with Big-GAN~\cite{brock2018large}, SA-GAN~\cite{han18self-attention}, and SN-GAN~\cite{miyato2018spectral}, PPCD-GAN reduces up to 5.2$\times$ (81\%) parameters (\#Par.) and 13.3$\times$ (92\%) MACs.
Although Tiny-GAN~\cite{chang2020tinygan} is more compact than our model, its IS, FID, and LPIPS are much poorer as seen above.

The out-performances of PPCD-GAN hold true when we compute the runtime (second per the batch of 64 images) on GPU and CPU for all the methods.
Indeed, the fourth block in Table~\ref{tab:quantitative_comparison} confirms that PPCD-GAN is (almost) faster than the other models.
We see that Tiny-GAN~\cite{chang2020tinygan} does not work properly on the CPU mode because it employs the depth-wise separable convolution, which is not optimized on CPU.

We also compare our method with other GANs compression methods.
Distill+NAS~\cite{li2020gan} indicates the model (SA-GAN~\cite{han18self-attention}) is first trained with knowledge distillation (i.e., learnable remapping~\cite{li2020gan}) and then compressed using NAS~\cite{elsken2019neural}.
Distill only~\cite{Aguinaldo2019Compressing} indicates the model is trained with knowledge distillation~\cite{hinton2015distilling} (we use the settings in the original paper~\cite{Aguinaldo2019Compressing}).
Table~\ref{tab:quantitative_compression} shows that other GANs compression methods~\cite{li2020gan,Aguinaldo2019Compressing} perform far worse than our method in both overall quality and diversity of generated images.
Distill only~\cite{Aguinaldo2019Compressing} is faster than ours, but its quality and diversity are worst.
This reveals that previous works~\cite{li2020gan,Aguinaldo2019Compressing} struggle in handling our novel task properly.
In contrast, our method successfully completes the task thanks to the proposal of simultaneously pruning and distillation.
Since Distill + NAS and Distill cannot produce reasonable results, we show their generated images in the supplementary.

\begin{figure}[tb]
	\centering	
	\includegraphics[width=1\linewidth]{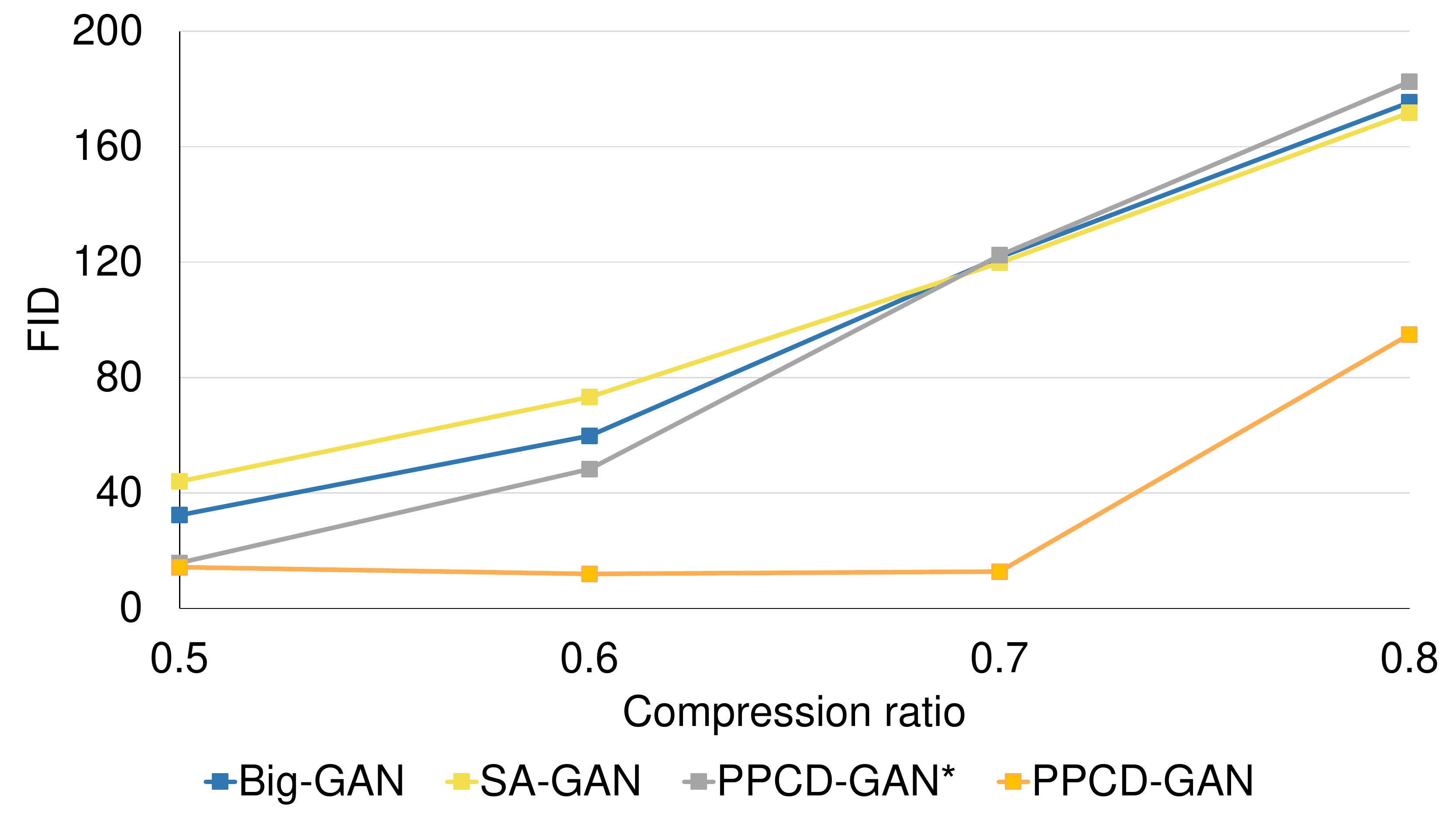}
		\vspace*{-1.5\baselineskip}	
	\caption{FID under various compression ratio thresholds ($\alpha=0.5, 0.6, 0.7, 0.8$).} 
	\label{fig:ratio}
	\vspace*{-0.3\baselineskip}
\end{figure}

Finally, to clearly illustrate the advance of PPCD-GAN against CNNs compression methods, 
we employed $\ell_1$ regularization~\cite{li2017pruning} to prune the network parameters of Big-GAN~\cite{brock2018large} and SA-GAN~\cite{han18self-attention} under various compression ratio thresholds (i.e., $\alpha=0.5, 0.6, 0.7, 0.8$), and then fine-tune the compressed model.
After that, for each compressed model, we generated images 10 times and computed the average of FID over the 10 times (Fig.~\ref{fig:ratio}).
We see that increasing the compression ratio leads to quickly degrading FIDs on Big-GAN~\cite{brock2018large} and SA-GAN~\cite{han18self-attention}.
In contrast, we see that PPCD-GAN sustainably generates high-quality images up to $\alpha=0.7$ even under different compression ratios while it fails in the case of $\alpha=0.8$.
It is obvious that the compression ratio of 0.8 makes the model size too small to handle a huge of classes properly.
These observations confirm much better compression capability of PPCD-GAN than the typical CNNs compression methods for GANs compression.

Through these various quantitative comparisons, we conclude that PPCD-GAN is superior to the others in both quality of generated images and complexity of the model.

\begin{table}[tb]
\centering
\caption{Quantitative comparison against other GANs compression methods (the best in \textcolor[rgb]{0,0,1}{blue}; the second best in \textcolor[rgb]{1,0,0}{red}).}
\vspace*{-0.75\baselineskip}
\label{tab:quantitative_compression}
\resizebox{\linewidth}{!}{
\begin{tabular}{l|ccccc}
\midrule
Models & IS $\uparrow$ &  FID $\downarrow$ & LPIPS $\uparrow$ & \#Par. $\downarrow$ & GPU $\downarrow$ \\
\midrule
Distill + NAS~\cite{li2020gan} & \textcolor[rgb]{1,0,0}{6.98} & \textcolor[rgb]{1,0,0}{217} & \textcolor[rgb]{1,0,0}{0.13} & 37M & 1.41 \\
Distill only~\cite{Aguinaldo2019Compressing}  & 3.34 & 272 & 0.09 & \textcolor[rgb]{0,0,1}{3.5M} & \textcolor[rgb]{0,0,1}{0.46} \\
\textbf{PPCD-GAN} ($\alpha=0.7$) & \textcolor[rgb]{0,0,1}{83.13} & \textcolor[rgb]{0,0,1}{12.76} & \textcolor[rgb]{0,0,1}{0.62} & \textcolor[rgb]{1,0,0}{13.6M} & \textcolor[rgb]{1,0,0}{1.19} \\
\hline
\end{tabular}
}
\vspace*{-0.2\baselineskip}
\end{table}

\subsection{Detailed analysis}

\begin{table}[tb]
\centering
\caption{Comparison of ablation models. We used $\alpha=0.5$ for all the models except for PPCD-GAN w/o $\mathcal{L}_{\rm PP}$ (the best in \textcolor[rgb]{0,0,1}{blue}; the second best in \textcolor[rgb]{1,0,0}{red}).}
\vspace*{-0.75\baselineskip}
\label{tab:ablation_study}
\begin{tabular}{l|ccc}
\midrule
Models & IS $\uparrow$ &  FID $\downarrow$  & \#Par. $\downarrow$ \\
\midrule
\textbf{PPCD-GAN} (complete) & \textcolor[rgb]{1,0,0}{81.47} & \textcolor[rgb]{1,0,0}{14.26} & \textcolor[rgb]{0,0,1}{25M} \\
PPCD-GAN w/o $\mathcal{L}_{\rm PP}$ & \textcolor[rgb]{0,0,1}{82.29} & \textcolor[rgb]{0,0,1}{13.02} & \textcolor[rgb]{1,0,0}{42M} \\
PPCD-GAN w/o $\mathcal{L}_{\rm CD}$  & 53.75 & 18.23 & \textcolor[rgb]{0,0,1}{25M} \\
\multicolumn{1}{c}{\dotfill} \\
PPCD-GAN* & 78.65 & 15.76 & \textcolor[rgb]{0,0,1}{25M} \\
\midrule
\end{tabular}
\vspace*{-1\baselineskip}
\end{table}

\begin{figure}[tb]
	\centering
	\includegraphics[width=0.95\linewidth]{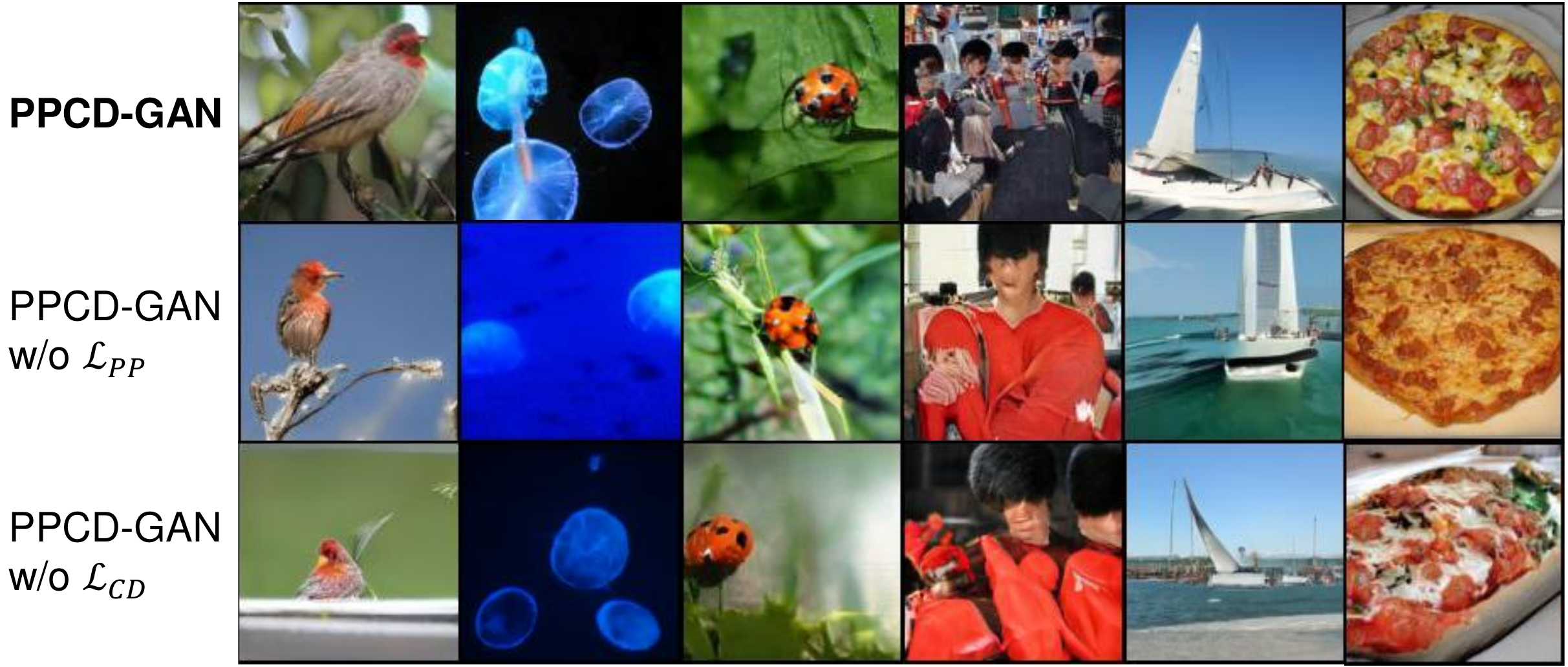}
		\vspace*{-0.5\baselineskip}
	\caption{Example results by the ablation models. From left to right, conditional classes are brambling, jellyfish, ladybug, bearskin, catamaran, and pizza.} 
	\label{fig:ablation}
\end{figure}

\noindent
\textbf{Ablation study.} 
We evaluated the necessity of employing PP-Res and class-aware distillation, see first three rows of Table~\ref{tab:ablation_study}. PPCD-GAN w/o $\mathcal{L}_{\rm PP}$ denotes the model dropping PP-Res (i.e., we trained SA-GAN~\cite{han18self-attention} with class-aware distillation), and PPCD-GAN w/o $\mathcal{L}_{\rm CD}$ denotes the model dropping class-aware distillation.
All the ablation models in this experiment are trained with $\alpha=0.5$ for fair comparison since the model not using class-aware distillation (i.e., PPCD-GAN w/o $\mathcal{L}_{\rm CD}$) does not work precisely with other compression ratios.
PPCD-GAN w/o $\mathcal{L}_{\rm PP}$ has no pruning (i.e., is not affected by $\alpha$).
We see that the ablation models suffer from the quality--complexity trade-off.
In particular, PPCD-GAN w/o $\mathcal{L}_{\rm PP}$ generates images with better IS and FID while PPCD-GAN w/o $\mathcal{L}_{\rm CD}$ is better in size of the model.
We conclude that the PP-Res and the class-aware distillation together bring gain the performance of our complete model.
Indeed, we can further compress PPCD-GAN (complete) (up to $\alpha=0.7$, Table~\ref{tab:quantitative_comparison}). 
It is worth noting that all the ablation models generated plausible images (Fig.~\ref{fig:ablation}).

\noindent
\textbf{Joint training.}
We evaluated the plausibility of jointly training the pruning and the class-aware distillation processes.
PPCD-GAN* denotes the two-step training model where we first pruned the model and then executed class-aware distillation.
The last row in Table~\ref{tab:ablation_study} shows the performances of PPCD-GAN* (with $\alpha=0.5$).
We see that although PPCD-GAN* (with $\alpha=0.5$) achieves comparable results with PPCD-GAN (complete), $\alpha=0.7$ cannot be used for PPCD-GAN* (see the ablation study above).
To further conduct fine-grained study on the limitation of PPCD-GAN*, we changed compression ratio $\alpha$ by 0.1 from 0.5 to 0.8, see Fig.~\ref{fig:ratio} (gray line).
We see that different from PPCD-GAN (complete), PPCD-GAN* degrades FIDs quickly and drastically as $\alpha$ increases.
We thus conclude that training both the processes jointly is plausible.
We note that training the two processes reversely is not applicable because pruning a trained model is risky as aforementioned.

\begin{figure}[tb]
	\centering
	\includegraphics[width=0.95\linewidth]{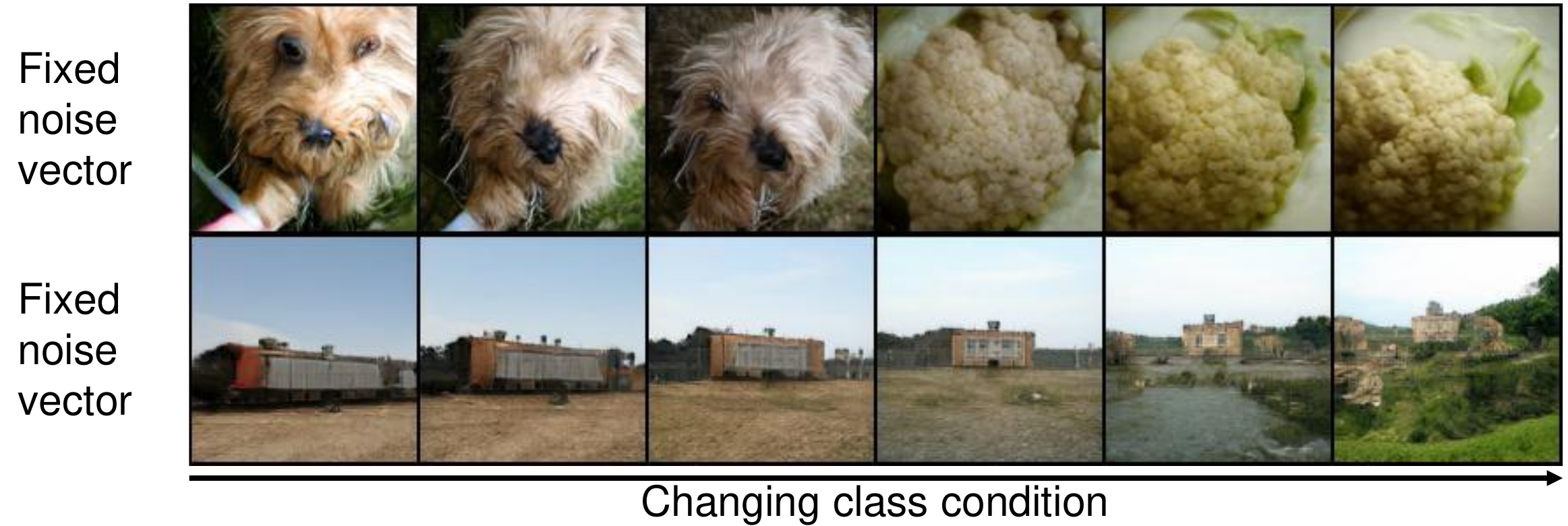}
			\vspace*{-0.5\baselineskip}
	\caption{Interpolated results by changing class condition while fixing noise vector.} 
	\label{fig:inter_y}
	\vspace*{-0.75\baselineskip}
\end{figure}

\begin{figure}[tb]
	\centering
	\includegraphics[width=0.95\linewidth]{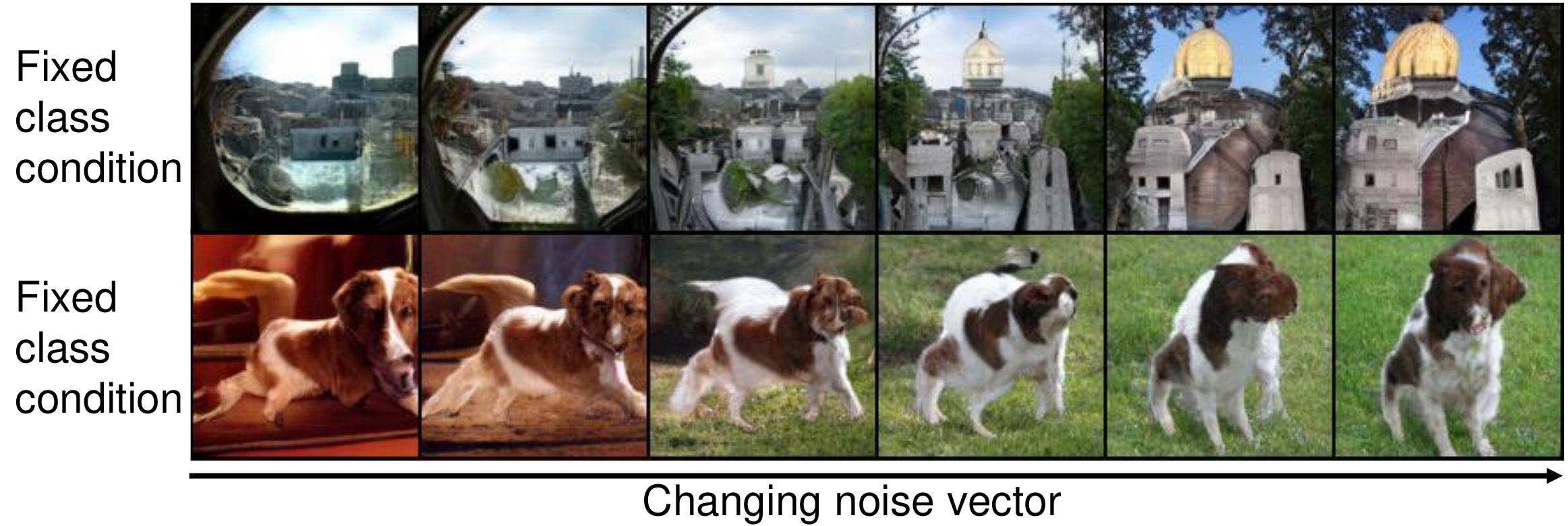}
		\vspace*{-0.5\baselineskip}	
	\caption{Interpolated results by fixing class condition while changing noise vector.} 
	\label{fig:inter_z}
\end{figure}

\noindent
\textbf{Interpolation.} We demonstrate the interpolation between the noise vector and the class condition vector in Figs~\ref{fig:inter_y} and~\ref{fig:inter_z}.
In Fig.~\ref{fig:inter_y}, we use a fixed noise vector while changing the class condition vector.
We observe that the viewpoint semantics are maintained consistently regardless of class.
In contrast, when we generate images with variant noise vectors and fixed class condition vector (Fig.~\ref{fig:inter_z}), the viewpoint as well as quantity are changed while the objects are unchanged.
These observations suggest that PPCD-GAN preserves the generalization ability even it is smaller in size.

\begin{figure}[tb]
	\centering
	\includegraphics[width=0.85\linewidth]{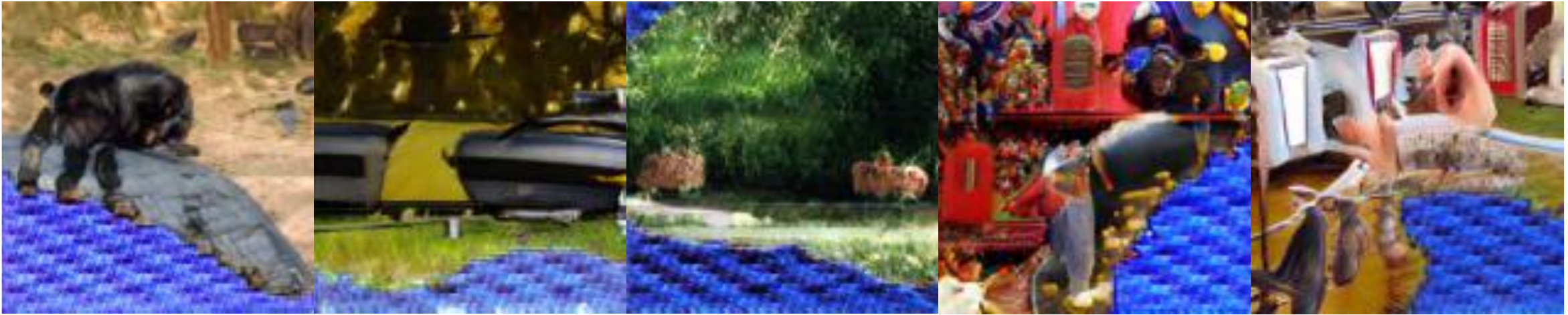}
			\vspace*{-0.5\baselineskip}
	\caption{Failure examples of generated images.} 
	\label{fig:failures}
	\vspace*{-1.25\baselineskip}
\end{figure}

\noindent
\textbf{Failure cases.} Though our method generates high-quality images, we find some failure cases when visually observing many generated images.
Some are generated with incomplete background (i.e., blue zone) (Fig.~\ref{fig:failures}).
Meanwhile, the main objects are apparent, leading to their not affecting the quantitative scores (IS, FID, and LPIPS).
Note that we observe that such failed images are rare.
One possible reason is that since PPCD-GAN pays more attention to ``important" parameters concerning main objects, parameters concerning the background may be primarily removed during training.
Detailed investigations are left for future work.

\section{Conclusion}

This paper presents PPCD-GAN for large-scale conditional GANs compression by introducing progressive pruning residual block (PP-Res) and class-aware distillation.
The PP-Res consists of the learnable mask layer and the transition layer, allowing us to progressively prune redundant network parameters during training.
Class-aware distillation stabilizes training through transferring immense knowledge from a teacher model to our model via attention maps.
We simultaneously train the pruning and distillation processes in the end-to-end fashion.
Our experiments on ImageNet show that PPCD-GAN is notably efficient in both parameter reduction and speed while achieving comparable image quality against state-of-the-arts.  

\noindent
\textbf{Acknowledgement.}
This work was supported by Institute of AI and Beyond of the University of Tokyo, and JSPS KAKENHI Grant Number JP19H04166.

{\small
\bibliographystyle{IEEEtran}
\bibliography{references}
}

\end{document}